\documentclass[letterpaper, 10 pt, conference]{ieeeconf}  % Comment this line out if you need a4paper

\IEEEoverridecommandlockouts                              % This command is only needed if 
\usepackage{times}
\usepackage{epsfig}
\usepackage{graphicx}
\usepackage{amsmath}
\usepackage{amssymb}
\usepackage{multirow}
\usepackage{amssymb}
\usepackage{pifont}
\usepackage{booktabs}
\usepackage{tabularx}
\def\name{MDSP}
\newcommand{\xmark}{\ding{55}}%

\title{\LARGE \bf
Multitask Network for Joint Object Detection, Semantic Segmentation and Human Pose Estimation in Vehicle Occupancy Monitoring
}

\author{Nikolas Ebert$^{1}$, Patrick Mangat$^{1}$ and Oliver Wasenm\"uller$^{1}$% <-this % stops a space
\thanks{$^{1}$Mannheim University of Applied Sciences, Germany}
\thanks{{\tt\small n.ebert@hs-mannheim.de}}
\thanks{{\tt\small p.mangat@hs-mannheim.de}}
\thanks{{\tt\small o.wasenmueller@hs-mannheim.de}}}

\begin{document}

\maketitle
\thispagestyle{empty}
\pagestyle{empty}

%%%%%%%%%%%%%%%%%%%%%%%%%%%%%%%%%%%%%%%%%%%%%%%%%%%%%%%%%%%%%%%%%%%%%%%%%%%%%%%%
\begin{abstract}

    In order to ensure safe autonomous driving, precise information about the conditions in and around the vehicle must be available. Accordingly, the monitoring of occupants and objects inside the vehicle is crucial. 
    In the state-of-the-art, single or multiple deep neural networks are used for either object recognition, semantic segmentation, or human pose estimation.
    In contrast, we propose our Multitask  Detection, Segmentation and Pose Estimation Network (\name{}) -- the first multitask network solving all these three tasks jointly in the area of occupancy monitoring. 
    Due to the shared architecture, memory and computing costs can be saved while achieving higher accuracy. 
    Furthermore, our architecture allows a flexible combination of the three  mentioned tasks during a simple end-to-end training.
    We perform comprehensive evaluations on the public datasets SVIRO and TiCaM in order to demonstrate the superior performance. 

\end{abstract}

%%%%%%%%%%%%%%%%%%%%%%%%%%%%%%%%%%%%%%%%%%%%%%%%%%%%%%%%%%%%%%%%%%%%%%%%%%%%%%%%
\section{Introduction}

% \begin{figure}[ht]%
%   \centering
%   \subfloat{\includegraphics[width=0.95\linewidth]{images/cover1.pdf}}%
%   \qquad
%   \subfloat{\includegraphics[width=0.95\linewidth]{images/cover2.pdf}}%
%   \qquad
%   \subfloat{\includegraphics[width=0.95\linewidth]{images/cover3.pdf}}%
%   \qquad
%   \caption{Exemplary results of our \name{} -- a multitask network for joint object detection, semantic segmentation and human pose estimation for occupancy monitoring on different datasets for the front as well as the back seats.}%
%   \vspace{-2mm}
% \label{fig:cover_img}
% \end{figure}

For autonomous driving, not only the environment outside the vehicle must be analyzed, but also the interior \cite{DiasDaCruz2021IV}. 
Information about the number and position of occupants as well as objects located inside the vehicle can contribute decisively to safety. 
For instance, airbags save the lives of the driver and passengers, but can also cause serious injury or even death to children who are in a rear-facing child seat in the front passenger seat \cite{nichols2005impact}.
In case of accidents, these kind of injuries can be prevented or at least limited via the controlled deployment of airbags \cite{farmer2003occupant}. 
In addition, interior monitoring can be used for daily situations such as a handover \cite{mccall2016towards} in autonomous driving.
Monitoring of the rear seats can also be of great interest, e.g.~in case of a forgotten baby in the vehicle  \cite{diewald2016rf}.

Interior monitoring is still a relatively new discipline in computer vision, partly because the recently released SVIRO \cite{DiasDaCruz2020SVIRO} and TiCaM \cite{katrolia2021ticam} datasets are the first to cover occupancy monitoring and provide data for segmentation, object detection and pose estimation.
So far the vast majority of existing approaches address these challenges using single task networks.
However, single task networks typically suffer from great computational effort and long inference times.

\begin{figure}[t]
\begin{center}
\includegraphics[width=0.9\linewidth]{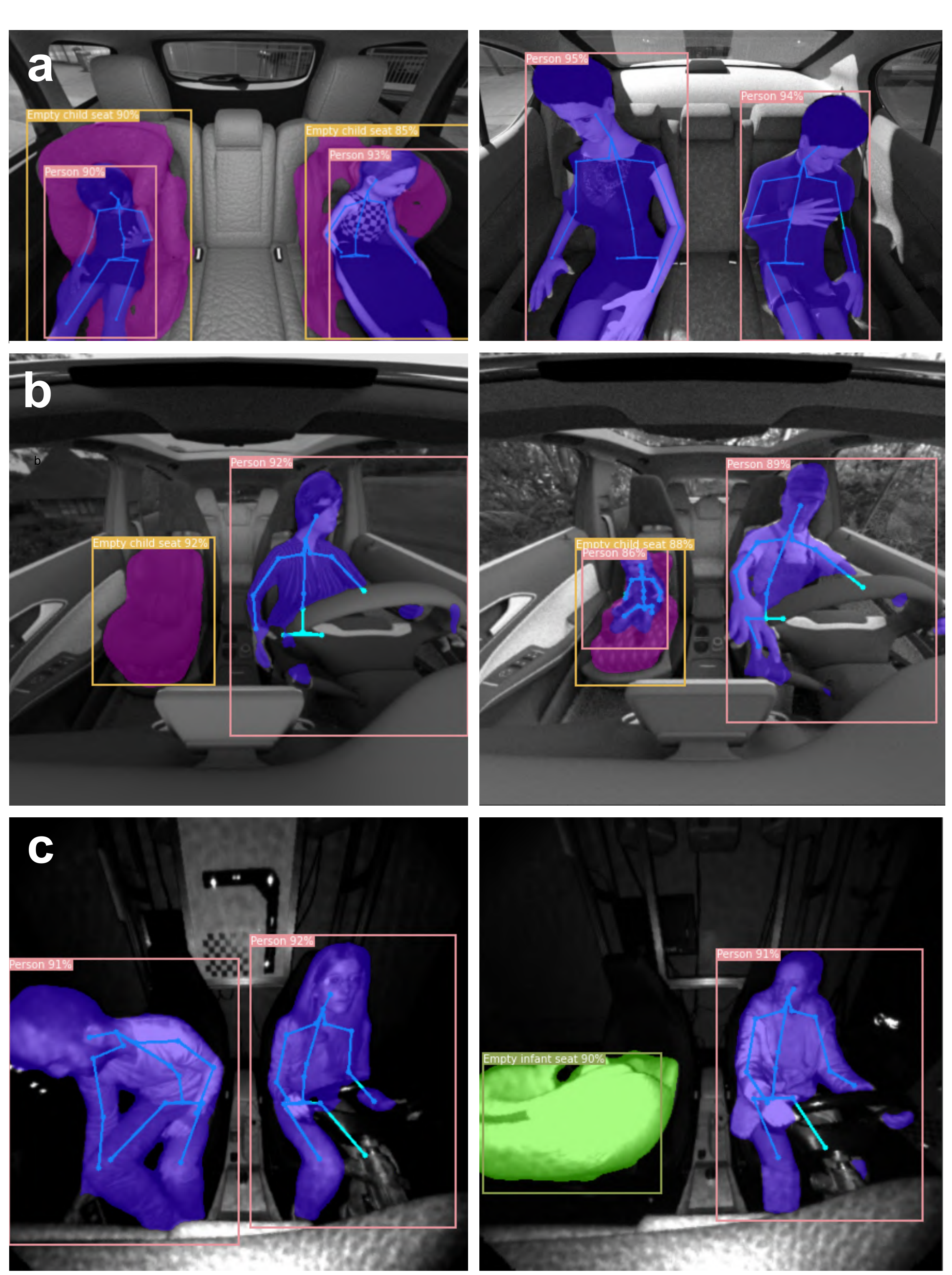}
\end{center}
   \caption{Exemplary results of our MDSP – a multitask network for joint
object detection, semantic segmentation and human pose estimation for
occupancy monitoring on (a) synthetic SVIRO [6] for the back as well as
the front seats on (b) synthetic and (c) real TiCAM [7].}\vspace{-2mm}
\label{fig:cover_img}
\end{figure}

In this paper, we present our multitask deep learning based approach \textit{\name{}} (\textbf{M}ultitask  \textbf{D}etection, \textbf{S}egmentation and \textbf{P}ose Estimation) for fully comprehensive monitoring of vehicle occupancy.
As illustrated in Figure \ref{fig:cover_img}, we perform a joint multi-class object detection and  semantic segmentation extended with a 2D multi-person pose estimation.
In contrast to state-of-the-art approaches using single task networks, we are the first in the field of occupancy monitoring using multitask learning to solve multiple tasks with a single network. 
On the one hand, this reduces the number of weights to save memory, and on the other hand, it can improve accuracy \cite{heuer2021multitask}.
Our novel architecture combines the heads of established networks \cite{cao2019openpose, lwRefinenet, redmon2018yolov3} using a shared backbone.
We evaluate our model through extensive experiments based on the two datasets SVIRO \cite{DiasDaCruz2020SVIRO} and TiCaM \cite{katrolia2021ticam} which are used for occupancy monitoring.
In these experiments, we consider front and rear seats in different vehicles using real and synthetic images.
Furthermore, intensive tests and comparisons are performed between the proposed multitask network and corresponding single task alternatives. 

%%%%%%%%%%%%%%%%%%%%%%%%%%%%%%%%%%%%%%%%%%%%%%%%%%%%%%%%%%%%%%%%%%%%%%%%%%%%%%%%%%%%

\section{Related Work}

\textbf{Multitask Learning.} In the last few years, multitask learning (MTL) \cite{he2017mask, kokkinos2017ubernet} has achieved great success in the field of computer vision. 
It is used to learn multiple tasks jointly, such as image classification in multiple domains \cite{rebuffi2017learning} or object detection and semantic segmentation \cite{dvornik2017blitznet}. 
Unlike classic single task learning (STL), which typically uses a backbone and a single network head to generate output, MTL generally employs multiple task-specific heads and shared backbone-layers \cite{Liu_2019_CVPR, teichmann2018multinet}. 
By maintaining shared feature representation even in deep network layers, we can profit from a symbiosis of the individual tasks \cite{caruana1997multitask}.
Network speed can also be increased and computing costs reduced. 

The combination of object detection and segmentation has already been addressed in various models.
For example, Mask R-CNN \cite{he2017mask} performs object detection and instance segmentation for multiple tasks. It can also be extended to include keypoint detection through additional one-hot masks for keypoint prediction. 
Similar to Mask R-CNN, MultiPoseNet \cite{kocabas2018multiposenet} is used for joint object detection, instance segmentation and human pose estimation, but it can only be applied to single class.
Kim et al. \cite{kim2019lightweight} present a multitask network to interior monitoring using lightweight mobilenets to determine head-pose and the state of the eyes and mouth.
To the best of our knowledge, Multitask CenterNet \cite{heuer2021multitask} is the only network that solves the problem of joint object detection and semantic segmentation for multiple classes augmented by human pose estimation. 
However, Multitask CenterNet does not explicitly address the problem of occupancy monitoring.
We close this gap by building the first multitask occupancy monitoring approach. 
It is based on established single task networks \cite{cao2019openpose, lwRefinenet, redmon2018yolov3} and performs tightly coupled joint object detection and semantic segmentation for multiple classes as well as human pose estimation.

\textbf{Single Task Learning.}
Since there are no competing MTL methods in occupancy monitoring, we examine single task approaches in more detail.
In recent years, deep learning approaches \cite{girshick2014rich, redmon2016you} have proven to be the way to go, outperforming classical methods in the active field of object detection. 
Most of the proposed methods can be divided into two general approaches, one stage \cite{redmon2016you} and two stage detectors \cite{Girshick_2015_ICCV, ren2015faster}. 
The advantage of the one stage method, in contrast to the two stage method, is that classification and localization are achieved in one step. 
%One the one hand this generally results in lower accuracy, but on the other hand to a significantly shorter interference time. 
On the one hand, this leads to a lower accuracy, but on the other hand, it also results in a significantly shorter interference time.
Aiming towards a fast multitask network with competitive accuracy, we choose a one stage approach for our \name{} by adapting the multi-scale detection which is the basis of YOLOv3 \cite{redmon2018yolov3}.

Similar to object detection, semantic segmentation also achieves great success using deep neural networks. 
The work of Long et al. \cite{long2015fully} on transforming an image classification network into a fully convolutional one is a crucial milestone in the field of semantic segmentation.  
The FCN architecture is also used in the field of occupancy monitoring by Dias da Cruz et al. \cite{DiasDaCruz2020SVIRO} to detect rear seat occupancy.
In our work, we choose some weight-optimized blocks of Light-Weight RefineNet \cite{lwRefinenet}, an encoder-decoder based structure, that are closely linked to the path of object detection. 

The third computer vision task to be considered is the 2D multi-person pose estimation, which can be divided into top-down and bottom-up approaches. 
In bottom-up approaches \cite{cao2017realtime}, the body joints are first detected in order to assign them to individual person instances.
Top-down approaches usually consist of a combination of person detector and a single-person pose estimation method. 
This leads to a high accuracy, but also to a higher inference time.
Guesdon et al. \cite{guesdon2021dripe} does further research with established state of the art top-down techniques for single-person driver pose estimation without an additional person detector.
The head of our human pose estimation branch is based on the method of Cao et al. \cite{cao2017realtime} which includes several stages for predicting pairwise relationships (partial affinity fields) between keypoints, followed by a final stage for predicting the keypoints themselves. 
This bottom-up approach allows us to quickly identify multiple persons at once.

\begin{figure*}[t]
\begin{center}
\includegraphics[width=.80\linewidth]{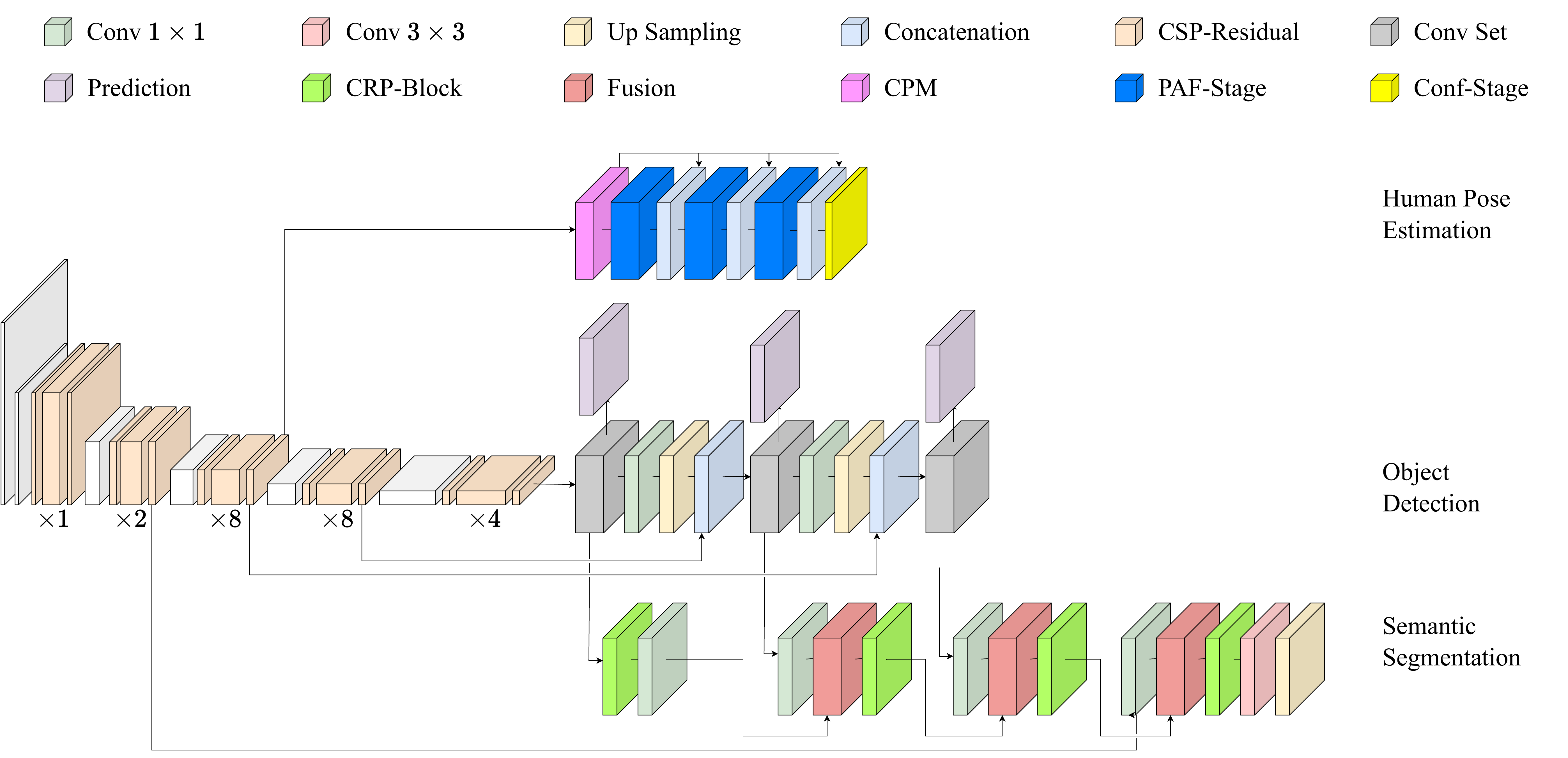}
\end{center}
   \caption{Illustration of the proposed network \name{}. The shared backbone CSPDarknet53 \cite{bochkovskiy2020yolov4, Wang_2020_CVPR_Workshops} is shown on the left. 
   The right side shows the head to human pose estimation, object detection and semantic segmentation from top to bottom. The head of semantic segmentation is closely connected to the branch of object detection.}
\label{fig:architecture}
\end{figure*}

By combining three real-time capable networks, we create with \name{} (see Figure \ref{fig:architecture}) a novel approach for fast joint object detection, semantic segmentation, and human pose estimation. 
Furthermore, \name{} reduces the number of trainable parameters and enables faster inferences compared to three single task networks.

\section{Method}
In the following section, we present our multitask network \name{} for comprehensive occupancy monitoring. 
Using multitask learning, in addition to reducing the weight of the network, an increase in accuracy should also be achieved through shared layers and joint training.  
The network is inspired by established single task approaches \cite{cao2019openpose, lwRefinenet, redmon2018yolov3} for fast joint object detection, semantic segmentation and human pose estimation.
Figure \ref{fig:architecture} shows the jointly end-to-end trainable encoder-decoder structure of our \name{}.
For feature extraction, we use the convolutional layers of CSPDarknet53 \cite{bochkovskiy2020yolov4, Wang_2020_CVPR_Workshops} pre-trained on ImageNet \cite{russakovsky2015imagenet}.
To solve the problem of object detection, we adapt the idea behind the head of YOLOv3 \cite{redmon2018yolov3} and fuse it with tightly linked chained residual pooling (CRP) and fusion blocks of the Light-Weight RefineNet \cite{lwRefinenet} for semantic segmentation.
The human pose estimation branch is loosely connected to the rest of the network, which allows us to provide more flexible training.  
This task is performed in several steps, following Cao et al. \cite{cao2017realtime} and including the prediction of partial affinity fields (PAFs) and a final confidence map.

% \begin{figure}[ht]
% \begin{center}
% \includegraphics[width=0.8\linewidth]{images/architecture_vertical.pdf}
% \end{center}
%   \caption{Illustration of the proposed network \name{}. The shared backbone CSPDarknet53 \cite{bochkovskiy2020yolov4, Wang_2020_CVPR_Workshops} is shown on the left. 
%   The right side shows the head to human pose estimation, object detection and semantic segmentation from top to bottom. The head of semantic segmentation is closely connected to the branch of object detection.}
% \label{fig:architecture}
% \end{figure}

For 2D bounding box detection, we use a multi-scale detection head as utilized in YOLOv3.
In this way, we can detect objects in three scales using high resolution features of shallow layers to better recognize objects of different sizes.
At the beginning we take backbone-features with a stride of 32 and make use of the first Conv Set to generate high level features for object detection.
Each  Conv Set consists of a sequence of three $1 \times 1$ convolution layers to halve the depth of the feature maps, interrupted by $3 \times 3$ layers for high level feature extraction and doubling the depth of the feature maps again.
The prediction layers follow the Conv Set and produce the outputs of the first scale. 
The output is based on the model of YOLOv3 in the form of an $N \times N \times [B \cdot (1 + 4 +C)]$ shaped tensor.
$N$ represents the spatial size of the tensor, $B$ is the number of anchor boxes and $C$ the number of classes. Moreover, to each anchor box we assign a confidence score for the objectness of the bounding box as well as four additional bounding box offsets.
We use k-means clustering to determine three anchors per scale.
In addition to the prediction layers, the feature maps of the Conv Set is also fed to another $1 \times 1$ layer to reduce the depth.
Afterwards, an upsampling by a factor of 2 followed by a concatenation with backbone features of the same resolution takes place to enhance context information.
The feature maps resulting from the concatenation are then passed to the next Conv Set and the process is repeated to generate the outputs of the next two resolutions.
Subsequently, a final non-maximum suppression is applied to the outputs.

The branch of semantic segmentation is highly connected to the layers of object detection.
The pixelwise classification of the input images as well as its refinement is done in four stages, following the example of RefineNet \cite{lin2017refinenet}. 
The input of the first segmentation-stage are the output feature maps of the first Conv Set in the object detection path.
This stage consists of a CRP followed by a single $1 \times 1$ convolutional layer to adjust the depth of the resulting features. 
The CRP aims to capture contextual background information from a large image region.
By sequencing multiple pooling layers, features are efficiently extracted under a stepwise increase of the receptive field and then fused together using a summation \cite{lin2017refinenet}.
Each CRP consists of an alternating chain of four max-pooling and convolutional layers. 
At the end, the outputs of each convolutional layer are fused in the manner of a residual network.
To reduce the number of parameters, we choose the weight-optimized CRP of Light-Weight RefineNet consisting of $5 \times 5$ pooling with a stride of 1 and $1 \times 1$ convolution layers.
The structure of stage 2 and 3 is similar to stage 1, but a fusion block is placed before the CRP.
Using $1 \times 1$ convolutional layers to adjust the depth and a subsequent upsampling, the output of the previous stage is fused with the output of the next Conv Set of object detection via summation and is subsequently activated with ReLU.
The resulting feature maps of these two stages are again obtained by CRP.
The fourth and final stage is structured like the previous stages, except that we do not use features from object detection. 
Instead, we directly use high resolution features from an early backbone layer. 
The omission of an additional Conv Set, which would only be used by the segmentation, is intended to save network parameters.
Finally, the resulting feature maps of the fourth stage are convolved with a $3 \times 3$ layer to produce feature maps with channels equal to the number of classes.
A final upsampling adjusts the spatial size of the output to the original image resolution.
The close connection between object detection and semantic segmentation described above enables joint feature learning. 
By learning semantic features directly in the path of object detection, a performance increase of both tasks should be achieved.

As can be seen in Figure \ref{fig:architecture}, the human pose estimation is only marginally connected to the rest of the network. 
We use backbone features with a stride of 8 for pose estimation, so only shallow encoder layers are shared by all three tasks.
At the beginning the aforementioned backbone features are processed by the convolutional pose machine (CPM) \cite{osokin2018real}, which consists of a sequence of ELU-activated convolutional layers to prepare features for the subsequent human pose estimation.
The CPM is followed by three stages to form PAFs \cite{cao2017realtime}, which are pairwise connections between keypoints.
Each stage consists of a series of five convolutional blocks and two $1 \times 1$ layers. 
A convolutional block in turn consists of a sequence of three $3 \times 3$ layers whose outputs are concatenated.
The input of each stage is a concatenation between the output of the previous stage and the features of the CPM. 
Compared to Cao et al. \cite{cao2017realtime}, we use only three stages to save parameters.
The final stage is created in the same way, but instead of PAFs, heatmaps are predicted for each keypoint.
Finally, heatmaps and PAFs of the last stage are resized to the original image resolution and grouping of keypoints takes place.
All layers in the pose estimation and object detection are activated via ReLU and batch-normalization is used.

We train all three tasks in parallel using the loss function:
\begin{equation}\label{eq:loss_total}
\mathcal{L} = \lambda_{dct}\mathcal{L}_{dct} + \lambda_{seg}\mathcal{L}_{seg} + \lambda_{pose}\mathcal{L}_{pose}
\end{equation}

The detection loss $\mathcal{L}_{dct}$ is a slightly modified version of YOLOv3-loss.
For the object and no-object losses, which penalize the network for finding no respectively wrong objects, we use the binary cross entropy loss.
Furthermore, we use the mean squared error to determine the loss of the bounding boxes, whereas the class error is calculated using the cross entropy loss.
The final detection loss $\mathcal{L}_{dct}$ is obtained via a weighted summation of the partial losses.
In the addition, the no-object-loss and the box-loss have a tenfold value, the remaining two components have a single value.
We calculate the loss $\mathcal{L}_{seg}$ of semantic segmentation using the cross entropy loss. 
Following the approach of Cao et al. \cite{cao2017realtime}, the L2 loss is used for each stage of the human pose estimation to determine the error of the PAFs and confidence scores.
The loss $\mathcal{L}_{pose}$ of the human pose estimation is determined via a single-weighted addition.
The final loss $\mathcal{L}$ defined in \eqref{eq:loss_total} is calculated using a weighted addition.
The weightings $\lambda_{dct}, \lambda_{seg}$ and $\lambda_{pose}$ are determined via the dynamic weight average \cite{Liu_2019_CVPR}.

\section{Evaluation}
%In the following, we show the performance of our \name{} through detailed evaluations.  
In the following, we evaluate the performance of our \name{}. 
For this purpose, the two datasets SVIRO \cite{DiasDaCruz2020SVIRO} and TiCaM \cite{katrolia2021ticam} for occupancy monitoring are used.  
For the training of our \name{}, we mainly use the SVIRO dataset.
SVIRO consists of 25,000 images, all synthetically generated and labeled for all required tasks.
In addition, SVIRO is currently the only dataset that provides ground truth data for these three tasks within interior, which is required for a joint training of our \name{}.
Each of these images shows a randomly generated interior scene from ten different vehicles. 
For the task of semantic segmentation and object detection, the shown instances are divided into the four classes \textit{Child seat}, \textit{Infant seat}, \textit{Person}, \textit{Everyday objects} and an additional fifth class \textit{Empty}, which represents the background. 
To investigate the network's ability to generalize to unseen interiors, we divide the images into three different sets for training and evaluation.
The first set \textit{All cars} contains the training data of all ten cars and is evaluated on the test data of these cars as well.
The second set \textit{5 Cars} contains five different vehicles and the third \textit{X5} only one car.
For the last two training sets, there are respective test sets that list all unused vehicles.

In addition to SVIRO, we performed other experiments on the TiCaM \cite{katrolia2021ticam} dataset.
This dataset consists of  synthetic and real \cite{feld2021dfki} annotated images for monitoring front seats and provides labels for object detection and semantic segmentation. 
For our experiments, the annotations provided by TiCaM are transformed into the five classes proposed by SVIRO. 

At the beginning of the training, we random initialize the network weights and perform a pre-training on MS COCO \cite{lin2014microsoft} over 50 epochs for all three tasks.
Subsequently, the network is trained another 50 epochs on SVIRO with a batch size of 4, regardless of the previously described selected training set. 
To achieve a low generalization-error \cite{zhou2020towards}, we pick stochastic gradient descent (SGD) with momentum over Adam. 
Furthermore, we choose a weight decay of $5 \cdot 10^{-5}$ and a momentum of 0.9. 
For SVIRO and TiCAM a learning rate of $lr = 1 \cdot 10^{-4}$ is chosen for the detectors, which is reduced  to $1 \cdot 10^{-5}$ for the second half of the training.
The learning rate of the backbone is $lr / 10$.
Typically, training and evaluation are performed on images with resolution $416 \times 416$.
Throughout the training, we use random rotation, mirroring, flipping, scaling, and noise as data augmentation.
The same hyperparameters are used for training on TiCaM.

\subsection{Ablation Study}

\textbf{Influence of multitask learning.} 
In the first experiment, the added value of multitask learning is investigated in more detail.
For this purpose, the network is trained as a single task as well as a multitask network with all possible task combinations.
If tasks are not learned, the learning rate is set to zero and no loss is computed.
The first step is a pre-training on MS COCO \cite{lin2014microsoft} followed by a fine-tuning on SVIRO \cite{DiasDaCruz2020SVIRO} using the training and test set \textit{5 Cars}.
The training is performed under the aforementioned conditions and the results are listed in Table \ref{tab:vgl_multi_single}.

\begin{table}[b]
\caption{Single and multitask networks trained for all classes of SVIRO \cite{DiasDaCruz2020SVIRO} on five different vehicles and evaluated on unseen interiors only.}
\centering
\label{tab:vgl_multi_single}
\begin{tabular}{cccccc}
\hline
\multicolumn{3}{c|}{\textbf{Task}}                               & \multicolumn{3}{c}{\textbf{Metric}}       \\
\textbf{Dct} & \textbf{Seg} & \multicolumn{1}{c|}{\textbf{Pose}} & \textbf{AP$_{dct}$} & \textbf{mIoU} & \textbf{AP$_{pose}$} \\ \hline
\multicolumn{6}{c}{\textit{\small{Single Task}}}  \\ \hline
\xmark             &              &                                    & 0.394                            & -                                  & -           \\
             & \xmark              &                                    & -                                & 0.453                              & -           \\
             &              & \xmark                                   & -                                & -                                  & 0.536       \\ \hline
\multicolumn{6}{c}{\textit{\small{Multitask}}}                                                                                                                 \\ \hline
\xmark             & \xmark             &                                    & 0.403                            & 0.507                              & -           \\
\xmark             &              & \xmark                                   & 0.442                            & -                                  & 0.591       \\
             &\xmark              & \xmark                                   & -                                & 0.457                              & 0.586       \\
\xmark             &\xmark              & \xmark                                   & 0.388                            & 0.566                              & 0.575       \\ \hline
\end{tabular}
\end{table}

\begin{figure}[t]
\begin{center}
\includegraphics[width=0.95\linewidth]{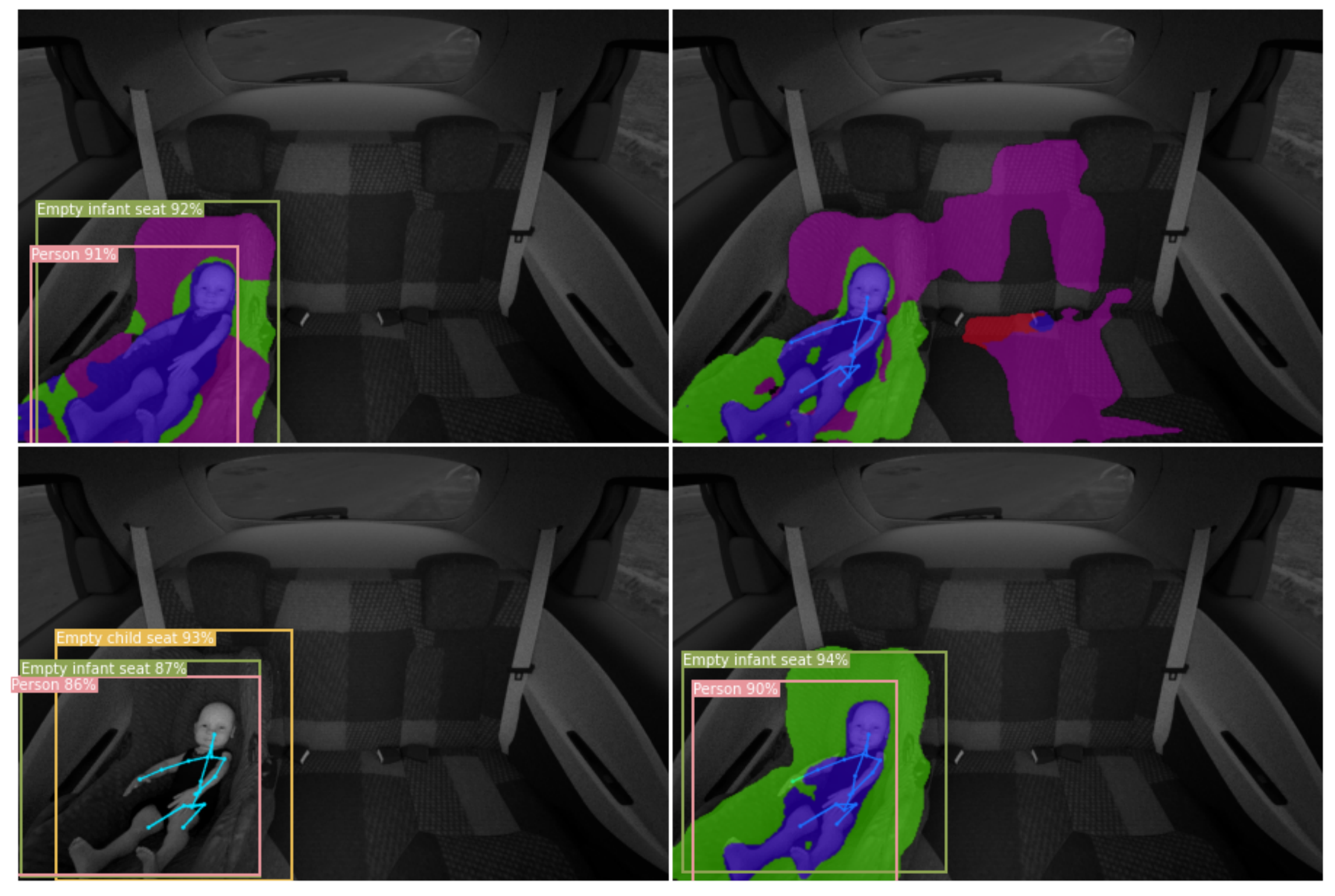}
\end{center}
\vspace{-3mm}
   \caption{Qualitative analysis of four possible task combinations of \name{}. Joint object detection and semantic segmentation (top left), joint object detection and human pose estimation (bottom left), joint semantic segmentation and human pose estimation (top right) and all three tasks together (bottom right). The analysis is visualized on a test sample of SVIRO \cite{DiasDaCruz2020SVIRO}.}
\label{fig:qual_multitask}\vspace{-3mm}
\end{figure}

Looking at the average precision AP$_{dct}$ of the object detection, we can see a strong benefit from the multitask learning.
Through the combination with the human pose estimation, the AP$_{dct}$ can be increased by almost 0.05 points to 0.442 from the original 0.394. 
The combination with semantic segmentation also leads to a slight increase in precision.
Only the multitask network consisting of all three tasks caused a slight negative transfer with an AP$_{dct}$ of 0.388.
In the case of semantic segmentation, a strong increase in performance can be seen in combination with object detection. 
This can be attributed to the close connection within the network architecture of the two tasks.
Coupled with object detection, the mIoU is increased by almost 0.05 points.
If the network is trained on all tasks together, we even find an increase in the mIoU of 0.11 points to 0.566.
The joint training with only human pose estimation does not lead to any appreciable increase in performance.
The human pose estimation benefits in all respects from the jointly trained backbone layers.
The AP$_{pose}$ increases by 0.04 to almost 0.06 points, depending on the combinations of tasks.
A qualitative example of multiple network combinations can be seen in Figure \ref{fig:qual_multitask}. 

\textbf{Generalization analysis.}
For a neural network to be used for occupancy monitoring, a high level of generalization is required.
Since only a limited number of different backgrounds (vehicle interiors), persons, everyday objects, as well as child and infant seats can be experienced during training, the network must be able to reliably classify and localize unknown objects in later use.
Thus, the generalization of the network to unseen vehicles is analyzed in more detail in the second experiment.
To this end, the network is trained on all three tasks jointly, using the aforementioned described training and test sets.
Furthermore, the influence of different numbers of classes on the network performance is investigated. 
The network is trained to two (\textit{Empty}, \textit{Person}), four (no \textit{Everyday Objects}) and all five classes.
All evaluation results are presented in Table \ref{tab:pose_sviro}.

With a look at the results using all classes, there is a minimal drop in the AP of the object detection on \textit{5 Cars} compared to \textit{All Cars}.
In contrast, a heavy drop in AP from 0.403 to 0.284 can be observed when using the training sets \textit{X5}. 
A similar behavior is found for semantic segmentation with a loss of nearly 0.1 (mIoU) and human pose estimation with one of 0.15 (AP) points.
This immense drop in performance indicates that the network uses vehicle-specific features such as texture as a feature to classify objects when training with a single vehicle.
If some different backgrounds are already seen during the training, it is much easier for the network to learn important vehicle-independent features.
A similar issue is also noted by Dias Da Cruz et al. \cite{DiasDaCruz2020SVIRO} in generating baseline results for SVIRO.

\begin{table}[t]
\centering
\scriptsize
\caption{Evaluation results on SVIRO \cite{DiasDaCruz2020SVIRO} using different number of images, backgrounds and classes for training. The evaluation is performed (if possible) on unseen vehicle types.}
\label{tab:pose_sviro}
\begin{tabular}{cccccccccc}
\hline
\multicolumn{1}{c|}{\multirow{2}{*}{\textbf{\begin{tabular}[c]{@{}c@{}}Train-\\ Set\end{tabular}}}} & \multicolumn{1}{c|}{\multirow{2}{*}{\textbf{\begin{tabular}[c]{@{}c@{}}No.\\ Class\end{tabular}}}} & \multicolumn{2}{c|}{\textbf{Detection}}                        & \multicolumn{2}{c|}{\textbf{Segmentation}}         & \multicolumn{2}{c}{\textbf{Pose}}          \\
\multicolumn{1}{c|}{}                                                                               & \multicolumn{1}{c|}{}                                                                                & \textbf{AP} & \multicolumn{1}{c|}{ \textbf{AP$_{50}$}} &\textbf{mIoU} & \multicolumn{1}{c|}{\textbf{Acc}} & \textbf{AP} & \textbf{AP$_{50}$} \\ \hline
\multirow{3}{*}{\begin{tabular}[c]{@{}c@{}}All\\ Cars\end{tabular}}                                 & 2                                                                                                    & 0.674        & 0.936                                           & 0.838         & 0.956                             & 0.593        & 0.785                      \\
                                                                                                    & 4                                                                                                    & 0.463        & 0.815                                          & 0.709         & 0.926                             & 0.616        & 0.824                    \\
                                                                                                    & 5                                                                                                    & 0.403        & 0.768                                          & 0.585         & 0.916                             & 0.588        & 0.792                      \\ \hline
\multirow{3}{*}{\begin{tabular}[c]{@{}c@{}}5\\ Cars\end{tabular}}                                   & 2                                                                                                    & 0.632        & 0.926                                          & 0.813         & 0.947                             & 0.566        & 0.797                       \\
                                                                                                    & 4                                                                                                    & 0.444        & 0.778                                          & 0.608         & 0.906                             & 0.595        & 0.803                      \\
                                                                                                    & 5                                                                                                    & 0.388              & 0.745                                                    &0.566               & 0.916                                  & 0.575             & 0.793                           \\ \hline
\multirow{3}{*}{X5}                                                                                 & 2                                                                                                    & 0.582        & 0.905                                       & 0.820         & 0.950                             & 0.457        & 0.658                  \\
                                                                                                    & 4                                                                                                    & 0.278            & 0.585                                                   & 0.564             & 0.876                                 & 0.453            & 0.648                          \\
                                                                                                    & 5                                                                                                    & 0.284            & 0.629                                                  & 0.482             & 0.883                                 & 0.437            & 0.638                           \\ \hline
\end{tabular}
\end{table}

\begin{table}[t]
\centering
\caption{Inference time of the single and multitask networks for object detection (Dct), semantic segmentation (Seg) and human pose estimation for processing $416 \times 416$ images. Inference time  with (Mode = \textit{CNN}) and without (Mode = \textit{Total}) post-processing is shown. Tests are done on a NVIDIA GeForce RTX 2060 Super (GPU$_1$), a NVIDIA Jetson AGX Xavier (GPU$_2$) and an AMD Ryzen 7 3800X (CPU). Trainable parameters (in million) are also given (last column).}
\label{tab:inf_time}
\footnotesize
\begin{tabular}{cccccccc}
\hline
\multicolumn{3}{c|}{\textbf{Task}}                                         & \multicolumn{1}{c|}{\multirow{2}{*}{\textbf{Mode}}} & \multicolumn{3}{c|}{\textbf{Time (ms)}}                                 & \multirow{2}{*}{\textbf{\begin{tabular}[c]{@{}c@{}}No.\\ Para\end{tabular}}} \\
\textbf{Dct}      & \textbf{Seg}      & \multicolumn{1}{c|}{\textbf{Pose}} & \multicolumn{1}{c|}{}                               & \textbf{GPU$_1$} & \textbf{GPU$_2$} & \multicolumn{1}{c|}{\textbf{CPU}} &                                                                               \\ \hline
\multicolumn{8}{c}{\textit{Single Task}}                                                                                                                                                                                                                                                   \\ \hline
\multirow{2}{*}{\xmark} & \multirow{2}{*}{} & \multirow{2}{*}{}                  & Total                                               & 22.4             & 129.4            & 243.6                             & \multirow{2}{*}{47.7}                                                         \\
                  &                   &                                    & CNN                                                 & 13.3             & 105.8            & 230.6                             &                                                                               \\ \hline
\multirow{2}{*}{} & \multirow{2}{*}{\xmark} & \multirow{2}{*}{}                  & Total                                               & 17.6             & 151.7            & 414.2                             & \multirow{2}{*}{29.7}                                                         \\
                  &                   &                                    & CNN                                                 & 17.3             & 138.2            & 389.9                             &                                                                               \\ \hline
\multirow{2}{*}{} & \multirow{2}{*}{} & \multirow{2}{*}{\xmark}                  & Total                                               & 81.6             & 386.8            & 448.9                             & \multirow{2}{*}{16.4}                                                         \\
                  &                   &                                    & CNN                                                 & 26.8             & 196.2            & 263.4                             &                                                                               \\ \hline
\multirow{2}{*}{\xmark} & \multirow{2}{*}{\xmark} & \multirow{2}{*}{}                  & Total                                               & 40.0             & 281.1            & 657.8                             & \multirow{2}{*}{77.4}                                                         \\
                  &                   &                                    & CNN                                                 & 30.6             & 244.0            & 620.7                             &                                                                               \\ \hline
\multirow{2}{*}{\xmark} & \multirow{2}{*}{\xmark} & \multirow{2}{*}{\xmark}                  & Total                                               & 121.6            & 667.9            & 1106.7                            & \multirow{2}{*}{93.8}                                                         \\
                  &                   &                                    & CNN                                                 & 57.4             & 440.2            & 883.9                             &                                                                               \\ \hline
\multicolumn{8}{c}{\textit{Multitask}}                                                                                                                                                                                                                                                     \\ \hline
\multirow{2}{*}{\xmark} & \multirow{2}{*}{\xmark} & \multirow{2}{*}{}                  & Total                                               & 27.7             & 175.2            & 439.4                             & \multirow{2}{*}{50.0}                                                         \\
                  &                   &                                    & CNN                                                 & 18.0             & 145.4            & 404.2                             &                                                                               \\ \hline
\multirow{2}{*}{\xmark} & \multirow{2}{*}{\xmark} & \multirow{2}{*}{\xmark}                  & Total                                               & 94.7             & 484.8            & 676.6                             & \multirow{2}{*}{64.7}                                                         \\
                  &                   &                                    & CNN                                                 & 34.7             & 252.4            & 612.7                             &                                                                               \\ \hline
\end{tabular}
\end{table}

% With regard to the evaluation results in terms of the number of classes, a decrease can be seen with increasing consideration of different classes.
% In Table \ref{tab:class_iou}, which shows the class-IoU of semantic segmentation and the class-AP of object detection, a strong deviation between the class of \textit{Person} (P) and the rest of the classes is evident.
% The everyday objects results, in particular, lag significantly behind the other classes. 
% This is further exacerbated by the use of fewer training images.

\begin{table*}[t]
\footnotesize
\centering
\caption{SVIRO \cite{DiasDaCruz2020SVIRO} benchmark analysis -- comparison of the leading methods on SVIRO for object detection, semantic segmentation and human pose estimation with our MDSP. Training is performed with BMW X5 data and evaluated for all unseen vehicles.}
\label{tab:benchmark}
\begin{tabular}{lccccccccccc}
\hline
\multicolumn{1}{l|}{\multirow{2}{*}{\textbf{Method}}} & \multicolumn{3}{c|}{\textbf{Task}}                                       & \multicolumn{3}{c|}{\textbf{Detection}}                                    & \multicolumn{2}{c|}{\textbf{Segmentation}}        & \multicolumn{3}{c}{\textbf{Pose}}                     \\
\multicolumn{1}{l|}{}                                 & \textbf{Dct} & \textbf{Seg}         & \multicolumn{1}{c|}{\textbf{Pose}} & \textbf{AP} & \textbf{AP$_{50}$} & \multicolumn{1}{c|}{\textbf{AP$_{75}$}} & \textbf{mIoU} & \multicolumn{1}{c|}{\textbf{Acc}} & \textbf{AP} & \textbf{AP$_{50}$} & \textbf{AP$_{75}$} \\ \hline
\multicolumn{12}{c}{\textit{Single Task Baseline}}                                                                                                                                                                                                                                                                        \\ \hline
Faster R-CNN \cite{ren2015faster}                    & \xmark        &                      &                                    & 0.349       & 0.653              & 0.314                                   & -             & -                                 & -           & -                  & -                  \\
YOLOv3 \cite{redmon2018yolov3}                       & \xmark        & \multicolumn{1}{l}{} &                                    & 0.354       & 0.691              & 0.294                                   & -             & -                                 & -           & -                  & -                  \\
FCN32s \cite{long2015fully}                          &              & \xmark                &                                    & -           & -                  & -                                       & 0.432         & -                                 & -           & -                  & -                  \\
Mask R-CNN \cite{he2017mask}                         &              &                      & \xmark                              & -           & -                  & -                                       & -             & -                                 & 0.150       & 0.617              & 0.006              \\ \hline
\multicolumn{12}{c}{\textit{Multitask}}                                                                                                                                                                                                                                                                                   \\ \hline
MDSP (ours)                                           & \xmark        & \xmark                &                                    & 0.394       & 0.699              & 0.395                                   & 0.533         & 0.906                             & -           & -                  & -                  \\
MDSP (ours)                                           & \xmark        & \xmark                & \xmark                              & 0.284       & 0.629              & 0.204                                   & 0.482         & 0.883                             & 0.437       & 0.638              & 0.473              \\ \hline
\end{tabular}
\end{table*}

\textbf{Inference time.}
The inference times of the multitask network in different task combinations are listed in Table \ref{tab:inf_time}.
Furthermore, the number of learnable parameters in the different network combinations is also shown.
The times are measured on a NVIDIA GeForce RTX 2060 Super (GPU$_1$), NVIDIA Jetson AGX Xavier (GPU$_2$) and an AMD Ryzen 7 3800X (CPU). 
In addition to runtime tests on a stationary GPU, tests on a mobile GPU and CPU are used to demonstrate the suitability of the network for use in vehicles for occupancy monitoring.
The resolution of the input images as well as the output of the segmentation are $416 \times 416$.
The \textit{CNN} mode represents the time needed by the network to generate the output in the form of a tensor.
The \textit{Total} mode, on the other hand, shows the time taken by the entire pipeline to generate an output. 
This mode includes all post-processing steps on both GPU and CPU.
The tested network has been trained on all classes and task using the set \textit{5 Cars}.

By using multitask learning, the number of learnable parameters of joint object detection (Dct) and semantic segmentation (Seg) can be reduced by over 35\% compared to the single task network (STN).
Using all three tasks, the number of parameters can be reduced by almost one third compared to the STN.
These lighter architectures reduce the inference time of object detection and semantic segmentation by almost 30\% compared to the STN.
Furthermore, these two tasks can be executed with a frame rate of 36 FPS in real time. 
All three tasks together in the form of a multitask network yield a 22\% lower inference time in direct comparison with the single task solutions.
It is even reduced by 27\% on less powerful mobile GPUs.

\begin{table}[b]
\caption{Evaluation results for single and multitask on TiCaM \cite{katrolia2021ticam}. The networks were trained and evaluated with synthetic (syn) and real data on the 5 SVIRO-Classes \cite{DiasDaCruz2020SVIRO}.}
\footnotesize
\centering
\label{tab:ticam}
\begin{tabular}{cccccccc}
\hline
\multicolumn{2}{c|}{\textbf{Task}}                    & \multicolumn{1}{c|}{\multirow{2}{*}{\textbf{Data}}} & \multicolumn{3}{c|}{\textbf{Detection}}                                    & \multicolumn{2}{c}{\textbf{Segmentation}} \\
\textbf{Dct}      & \multicolumn{1}{c|}{\textbf{Seg}} & \multicolumn{1}{c|}{}                               & \textbf{AP} & \textbf{AP$_{50}$} & \multicolumn{1}{c|}{\textbf{AP$_{75}$}} & \textbf{mIoU}        & \textbf{Acc}       \\ \hline
\multicolumn{8}{c}{\textit{Single Task}}                                                                                                                                                                                             \\ \hline
\multirow{2}{*}{\xmark} & \multirow{2}{*}{}                 & syn                                                 & 0.694       & 0.990              & 0.808                                   & -                    & -                  \\
                  &                                   & real                                                & 0.325       & 0.413              & 0.494                                   & -                    & -                  \\ \hline
\multirow{2}{*}{} & \multirow{2}{*}{\xmark}                 & syn                                                 & -           & -                  & -                                       & 0.815                & 0.979              \\
                  &                                   & real                                                & -           & -                  & -                                       & 0.837                & 0.977              \\ \hline
\multicolumn{8}{c}{\textit{Multitask}}                                                                                                                                                                                               \\ \hline
\multirow{2}{*}{\xmark} & \multirow{2}{*}{\xmark}                 & syn                                                 & 0.701       & 0.993              & 0.817                                   & 0.807                & 0.974              \\
                  &                                   & real                                                & 0.349       & 0.494              & 0.440                                   & 0.872                & 0.979              \\ \hline
\end{tabular}
\end{table}

\textbf{Experiments on TiCaM.}
In addition to SVIRO, we are also running some experiments with TiCaM \cite{katrolia2021ticam}, which only provides ground truth data for object detection and semantic segmentation.
These tests are performed using synthetic as well as real images, as can be seen in Figure \ref{fig:cover_img}.
For this purpose, we generate a validation set of 500 random images from the training data for each dataset.
The training is performed on networks pre-trained with SVIRO over 20 epochs and in case of divergence it is stopped early.
For all learning rates, a value of $10^{-5}$ is chosen excluding the human pose estimation layers, which are frozen.
The evaluation for single task object detection and semantic segmentation, as well as the multitask combinations of these two tasks are listed in Table \ref{tab:ticam}.
As can be seen, especially when using real images, the accuracy in terms of AP and mIoU is enhanced by multitask learning.
The AP increases by 0.024, the mIoU even by 0.035 points.
Using synthetic images, we have a minor increase in AP of object detection and a slightly smaller mIoU of segmentation.

\subsection{Benchmark Evaluation}
\begin{figure*}[ht]
\begin{center}
\includegraphics[width=1\linewidth]{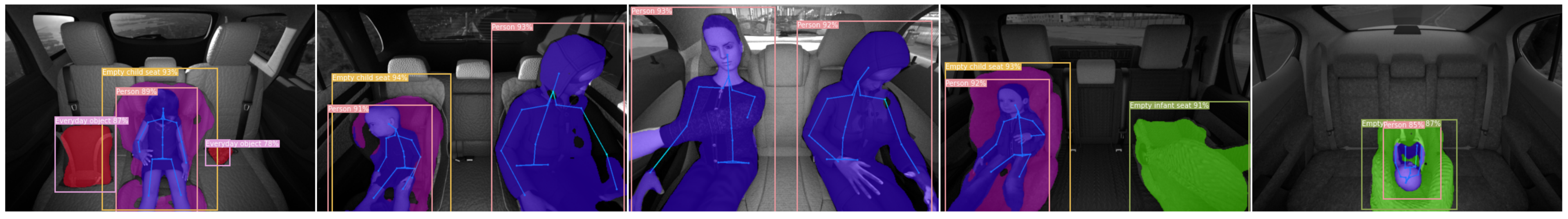}
\end{center}
\vspace{-3mm}
  \caption{\name{} applied to diverse example scenarios on SVIRO \cite{DiasDaCruz2020SVIRO} in different vehicles.}
\label{fig:example}
\end{figure*}

In order to compare the performance of our \name{} with the benchmark leading methods, we perform a training with the BMW X5 data and evaluate it with all unseen vehicles included in SVIRO \cite{DiasDaCruz2020SVIRO}.
At the current date, the benchmark leading methods are Faster R-CNN \cite{ren2015faster}, FCN32s \cite{long2015fully}, and Mask R-CNN \cite{he2017mask}, which are trained on the dataset by the SVIRO-Team \cite{DiasDaCruz2020SVIRO} to generate a baseline.
In addition, we create a further baseline in the area of one stage object detection with YOLOv3 \cite{redmon2018yolov3}.
The networks are trained on the BMW X5 and evaluated on all unseen vehicles, too.
As can be seen from Table \ref{tab:benchmark}, we outperform the single task baseline by over 0.04 points in the object detection domain with our \name{} (Seg + Dct). Looking at the AP$_{75}$ in particular, we even achieve an increase of almost 0.1 points.
Furthermore, the mIoU of the semantic segmentation is also exceeded by more than 0.1 points.
The poorer performance of the network consisting of all three tasks can be attributed to the limitation of training data to images containing persons due to inclusion of human pose estimation.
However, with our network, we can significantly increase the AP of human pose estimation, especially the AP$_{75}$. 
Figure \ref{fig:example} shows a number of inference examples from different vehicles of our \name{} on SVIRO.

\section{Conclusion}
In this paper, we presented our multitask network \name{} for joint object detection, semantic segmentation and human pose estimation.
Our \name{} consists of a combination of modified heads from three established single task networks \cite{cao2019openpose, lwRefinenet, redmon2018yolov3}, which all share a common backbone.
Despite the close connection, all tasks can be trained independently of each other, which admits a flexible combination of them.
By solving several tasks jointly, an increase in performance can be achieved. 
In addition, due to the shared network layers, significant savings can be gained in terms of the number of learnable parameters.
This in turn leads to a reduction in inference time of the network.
The aforementioned aspects  make \name{} ideal for occupancy monitoring. 
Using multi-class semantic segmentation and object detection extended with human pose estimation, a complete coverage of the situation in vehicles can be achieved.
With the training on several datasets for occupancy monitoring, the suitability for mastering this task is repeatedly demonstrated.

%%%%%%%%%%%%%%%%%%%%%%%%%%%%%%%%%%%%%%%%%%%%%%%%%%%%%%%%%%%%%%%%%%%%%%%%%%%%%%%%%
\addtolength{\textheight}{-12cm}   % This command serves to balance the column lengths

%%%%%%%%%%%%%%%%%%%%%%%%%%%%%%%%%%%%%%%%%%%%%%%%%%%%%%%%%%%%%%%%%%%%%%%%%%%%%%%%

\section*{ACKNOWLEDGMENT}
This work was funded by the Karl V\"olker Foundation. % in the project "KI-Fusion".

%\section{References}
\bibliographystyle{IEEEtran}
\bibliography{IEEEabrv,literatur}

\begin{thebibliography}{10}
\providecommand{\url}[1]{#1}
\csname url@samestyle\endcsname
\providecommand{\newblock}{\relax}
\providecommand{\bibinfo}[2]{#2}
\providecommand{\BIBentrySTDinterwordspacing}{\spaceskip=0pt\relax}
\providecommand{\BIBentryALTinterwordstretchfactor}{4}
\providecommand{\BIBentryALTinterwordspacing}{\spaceskip=\fontdimen2\font plus
\BIBentryALTinterwordstretchfactor\fontdimen3\font minus
  \fontdimen4\font\relax}
\providecommand{\BIBforeignlanguage}[2]{{%
\expandafter\ifx\csname l@#1\endcsname\relax
\typeout{** WARNING: IEEEtran.bst: No hyphenation pattern has been}%
\typeout{** loaded for the language `#1'. Using the pattern for}%
\typeout{** the default language instead.}%
\else
\language=\csname l@#1\endcsname
\fi
#2}}
\providecommand{\BIBdecl}{\relax}
\BIBdecl

\bibitem{DiasDaCruz2020SVIRO}
S.~{Dias Da Cruz}, O.~Wasenm\"uller, H.-P. Beise, T.~Stifter, and D.~Stricker,
  ``Sviro: Synthetic vehicle interior rear seat occupancy dataset and
  benchmark,'' in \emph{Winter Conference on Applications of Computer Vision
  (WACV)}, 2020.

\bibitem{katrolia2021ticam}
J.~S. Katrolia, B.~Mirbach, A.~El-Sherif, H.~Feld, J.~Rambach, and D.~Stricker,
  ``Ticam: A time-of-flight in-car cabin monitoring dataset,'' in \emph{The
  British Machine Vision Conference (BMVC)}, 2021.

\bibitem{DiasDaCruz2021IV}
S.~{Dias Da Cruz}, B.~Taetz, O.~Wasenm\"uller, T.~Stifter, and D.~Stricker,
  ``Autoencoder based inter-vehicle generalization for in-cabin occupant
  classification,'' in \emph{Intelligent Vehicles Symposium (IV)}, 2021.

\bibitem{nichols2005impact}
J.~L. Nichols, D.~Glassbrenner, and R.~P. Compton, ``The impact of a nationwide
  effort to reduce airbag-related deaths among children: an examination of
  fatality trends among younger and older age groups,'' \emph{Journal of safety
  research}, 2005.

\bibitem{farmer2003occupant}
M.~E. Farmer and A.~K. Jain, ``Occupant classification system for automotive
  airbag suppression,'' in \emph{Conference on Computer Vision and Pattern
  Recognition (CVPR)}, 2003.

\bibitem{mccall2016towards}
R.~McCall, F.~McGee, A.~Meschtscherjakov, N.~Louveton, and T.~Engel, ``Towards
  a taxonomy of autonomous vehicle handover situations,'' in
  \emph{International Conference on Automotive User Interfaces and Interactive
  Vehicular Applications (AutoUI)}, 2016.

\bibitem{diewald2016rf}
A.~R. Diewald, J.~Landwehr, D.~Tatarinov, P.~D.~M. Cola, C.~Watgen, C.~Mica,
  M.~Lu-Dac, P.~Larsen, O.~Gomez, and T.~Goniva, ``Rf-based child occupation
  detection in the vehicle interior,'' in \emph{International Radar Symposium
  (IRS)}, 2016.

\bibitem{heuer2021multitask}
F.~Heuer, S.~Mantowsky, S.~Bukhari, and G.~Schneider, ``Multitask-centernet
  (mcn): Efficient and diverse multitask learning using an anchor free
  approach,'' in \emph{International Conference on Computer Vision (ICCV)},
  2021.

\bibitem{cao2019openpose}
Z.~Cao, G.~Hidalgo, T.~Simon, S.-E. Wei, and Y.~Sheikh, ``Openpose: realtime
  multi-person 2d pose estimation using part affinity fields,''
  \emph{Tnsactions on Pattern Analysis and Machine Intelligence (PAMI)}, 2019.

\bibitem{lwRefinenet}
V.~Nekrasov, C.~Shen, and I.~D. Reid, ``Light-weight refinenet for real-time
  semantic segmentation,'' in \emph{The British Machine Vision Conference
  (BMVC)}, 2018.

\bibitem{redmon2018yolov3}
J.~Redmon and A.~Farhadi, ``Yolov3: An incremental improvement,'' \emph{arXiv
  preprint arXiv:1804.02767}, 2018.

\bibitem{he2017mask}
K.~He, G.~Gkioxari, P.~Doll{\'a}r, and R.~Girshick, ``Mask r-cnn,'' in
  \emph{International Conference on Computer Vision (ICCV)}, 2017.

\bibitem{kokkinos2017ubernet}
I.~Kokkinos, ``Ubernet: Training a universal convolutional neural network for
  low-, mid-, and high-level vision using diverse datasets and limited
  memory,'' in \emph{Conference on Computer Vision and Pattern Recognition
  (CVPR)}, 2017.

\bibitem{rebuffi2017learning}
S.-A. Rebuffi, H.~Bilen, and A.~Vedaldi, ``Learning multiple visual domains
  with residual adapters,'' \emph{arXiv preprint arXiv:1705.08045}, 2017.

\bibitem{dvornik2017blitznet}
N.~Dvornik, K.~Shmelkov, J.~Mairal, and C.~Schmid, ``Blitznet: A real-time deep
  network for scene understanding,'' in \emph{International Conference on
  Computer Vision (ICCV)}, 2017.

\bibitem{Liu_2019_CVPR}
S.~Liu, E.~Johns, and A.~J. Davison, ``End-to-end multi-task learning with
  attention,'' in \emph{Conference on Computer Vision and Pattern Recognition
  (CVPR)}, 2019.

\bibitem{teichmann2018multinet}
M.~Teichmann, M.~Weber, M.~Zoellner, R.~Cipolla, and R.~Urtasun, ``Multinet:
  Real-time joint semantic reasoning for autonomous driving,'' in
  \emph{Intelligent Vehicles Symposium (IV)}, 2018.

\bibitem{caruana1997multitask}
R.~Caruana, ``Multitask learning,'' \emph{Machine learning}, 1997.

\bibitem{kocabas2018multiposenet}
M.~Kocabas, S.~Karagoz, and E.~Akbas, ``Multiposenet: Fast multi-person pose
  estimation using pose residual network,'' in \emph{European Conference on
  Computer Vision (ECCV)}, 2018.

\bibitem{kim2019lightweight}
W.~Kim, W.-S. Jung, and H.~K. Choi, ``Lightweight driver monitoring system
  based on multi-task mobilenets,'' \emph{Multidisciplinary Digital Publishing
  Institute (MDPI)}, 2019.

\bibitem{girshick2014rich}
R.~Girshick, J.~Donahue, T.~Darrell, and J.~Malik, ``Rich feature hierarchies
  for accurate object detection and semantic segmentation,'' in
  \emph{Conference on Computer Vision and Pattern Recognition (CVPR)}, 2014.

\bibitem{redmon2016you}
J.~Redmon, S.~Divvala, R.~Girshick, and A.~Farhadi, ``You only look once:
  Unified, real-time object detection,'' in \emph{Conference on Computer Vision
  and Pattern Recognition (CVPR)}, 2016.

\bibitem{Girshick_2015_ICCV}
R.~Girshick, ``Fast r-cnn,'' in \emph{International Conference on Computer
  Vision (ICCV)}, 2015.

\bibitem{ren2015faster}
S.~Ren, K.~He, R.~Girshick, and J.~Sun, ``Faster r-cnn: Towards real-time
  object detection with region proposal networks,'' \emph{Neural Information
  Processing Systems (NIPS)}, 2015.

\bibitem{long2015fully}
J.~Long, E.~Shelhamer, and T.~Darrell, ``Fully convolutional nzetworks for
  semantic segmentation,'' in \emph{Conference on Computer Vision and Pattern
  Recognition (CVPR)}, 2015.

\bibitem{cao2017realtime}
Z.~Cao, T.~Simon, S.-E. Wei, and Y.~Sheikh, ``Realtime multi-person 2d pose
  estimation using part affinity fields,'' in \emph{Conference on Computer
  Vision and Pattern Recognition (CVPR)}, 2017.

\bibitem{guesdon2021dripe}
R.~Guesdon, C.~Crispim-Junior, and L.~Tougne, ``Dripe: A dataset for human pose
  estimation in real-world driving settings,'' in \emph{International
  Conference on Computer Vision (ICCV)}, 2021.

\bibitem{bochkovskiy2020yolov4}
A.~Bochkovskiy, C.-Y. Wang, and H.-Y.~M. Liao, ``Yolov4: Optimal speed and
  accuracy of object detection,'' \emph{arXiv preprint arXiv:2004.10934}, 2020.

\bibitem{Wang_2020_CVPR_Workshops}
C.-Y. Wang, H.-Y.~M. Liao, Y.-H. Wu, P.-Y. Chen, J.-W. Hsieh, and I.-H. Yeh,
  ``Cspnet: A new backbone that can enhance learning capability of cnn,'' in
  \emph{Conference on Computer Vision and Pattern Recognition (CVPR)}, 2020.

\bibitem{russakovsky2015imagenet}
O.~Russakovsky, J.~Deng, H.~Su, J.~Krause, S.~Satheesh, S.~Ma, Z.~Huang,
  A.~Karpathy, A.~Khosla, M.~Bernstein \emph{et~al.}, ``Imagenet large scale
  visual recognition challenge,'' \emph{International Journal of Computer
  Vision (IJCV)}, 2015.

\bibitem{lin2017refinenet}
G.~Lin, A.~Milan, C.~Shen, and I.~Reid, ``Refinenet: Multi-path refinement
  networks for high-resolution semantic segmentation,'' in \emph{Conference on
  Computer Vision and Pattern Recognition (CVPR)}, 2017.

\bibitem{osokin2018real}
D.~Osokin, ``Real-time 2d multi-person pose estimation on cpu: Lightweight
  openpose,'' \emph{arXiv preprint arXiv:1811.12004}, 2018.

\bibitem{feld2021dfki}
H.~Feld, B.~Mirbach, J.~Katrolia, M.~Selim, O.~Wasenm{\"u}ller, and
  D.~Stricker, ``Dfki cabin simulator: A test platform for visual in-cabin
  monitoring functions,'' in \emph{Commercial Vehicle Technology (CVT)}, 2021.

\bibitem{lin2014microsoft}
T.-Y. Lin, M.~Maire, S.~Belongie, J.~Hays, P.~Perona, D.~Ramanan,
  P.~Doll{\'a}r, and C.~L. Zitnick, ``Microsoft coco: Common objects in
  context,'' in \emph{European Conference on Computer Vision (ECCV)}, 2014.

\bibitem{zhou2020towards}
P.~Zhou, J.~Feng, C.~Ma, C.~Xiong, S.~Hoi \emph{et~al.}, ``Towards
  theoretically understanding why sgd generalizes better than adam in deep
  learning,'' \emph{arXiv preprint arXiv:2010.05627}, 2020.

\end{thebibliography}
%\printbibliography

\end{document}